\documentclass[conference,compsoc]{IEEEtran}
\IEEEoverridecommandlockouts
\usepackage{cite}
\usepackage{amsmath,amssymb,amsfonts}
\usepackage{algorithmic}
\usepackage{graphicx}
\usepackage{textcomp}
\usepackage{xcolor}
\usepackage{multirow}
\usepackage{xspace}
\usepackage{hyperref}
\usepackage{url}
\hypersetup{hidelinks}

\newcommand{\sys}{\ensuremath{\texttt{FSRDP}}}
\newcommand{\rdp}{\ensuremath{\texttt{RDP}}}
\newcommand{\nonoise}{\ensuremath{\texttt{Non-Private}}}

\newcommand{\oursys}{\ensuremath{\texttt{FLIP}}}

\def\BibTeX{{\rm B\kern-.05em{\sc i\kern-.025em b}\kern-.08em
    T\kern-.1667em\lower.7ex\hbox{E}\kern-.125emX}}
\begin{document}

\title{An Interactive Framework for Implementing Privacy-Preserving Federated Learning: Experiments on Large Language Models}

\makeatletter
\newcommand{\linebreakand}{%
  \end{@IEEEauthorhalign}
  \hfill\mbox{}\par
  \mbox{}\hfill\begin{@IEEEauthorhalign}
}
\makeatother
    \author{
    \IEEEauthorblockN{Kasra Ahmadi}
    \IEEEauthorblockA{ 
    \textit{University of South Florida}\\
    Tampa, FL 33620 \\
    ahmadi1@usf.edu}
    \and
    \IEEEauthorblockN{Rouzbeh Behnia}
    \IEEEauthorblockA{ 
    \textit{University of South Florida}\\
    Tampa, FL 33620 \\
    behnia@usf.edu}
    \and
    \IEEEauthorblockN{Reza Ebrahimi}
    \IEEEauthorblockA{ 
    \textit{University of South Florida}\\
    Tampa, FL 33620 \\
    ebrahimim@usf.edu}
    \linebreakand
    \IEEEauthorblockN{Mehran Mozaffari Kermani}
    \IEEEauthorblockA{ 
    \textit{University of South Florida}\\
    Tampa, FL 33620 \\
    mehran2@usf.edu}
    \and
    \IEEEauthorblockN{Jeremiah Birrell}
    \IEEEauthorblockA{ 
    \textit{Texas State University}\\
    San Marcos, TX 78666 \\
    jbirrell@txstate.edu}
    \and
    \IEEEauthorblockN{Jason Pacheco}
    \IEEEauthorblockA{
    \textit{University of Arizona}\\
    Tucson, AZ 85721 \\
    pachecoj@cs.arizona.edu}
    \and
    \IEEEauthorblockN{Attila A. Yavuz}
    \IEEEauthorblockA{
    \textit{University of South Florida}\\
    Tampa, FL 33620 \\
    attilaayavuz@usf.edu}
    }
    
\maketitle
\begin{abstract}

Federated learning (FL) enhances privacy by keeping user data on local devices. 
However, emerging attacks have demonstrated that the updates shared by users during training can reveal significant information about their data.
%
%
Differential Privacy (DP) is considered the gold standard for safeguarding user data.
However, DP  guarantees are highly conservative, providing worst-case privacy guarantees. 
This can result in overestimating privacy needs, which may compromise the model's accuracy. 
Additionally, interpretations of these privacy guarantees have proven to be challenging in different contexts.
This is further exacerbated when other factors, such as the number of training iterations, data distribution, and specific application requirements, can add further complexity to this problem.
In this work, we proposed a framework that integrates a human entity as a privacy practitioner to determine an optimal trade-off between the model's privacy and utility.
Our framework is the first to address the variable memory requirement of existing DP methods in FL settings, where resource-limited devices (e.g., cell phones) can participate. To support such settings, we adopt a recent DP method with fixed memory usage to ensure scalable private FL. 

 We evaluated our proposed framework by fine-tuning a BERT-based LLM model using the GLUE dataset (a common approach in literature), leveraging the new accountant, and employing diverse data partitioning strategies to mimic real-world conditions.
As a result, we achieved stable memory usage, with an average accuracy reduction of 1.33\% for \(\epsilon = 10\) and 1.9\% for \(\epsilon = 6\), when compared to the state-of-the-art DP~accountant which does not support fixed memory usage.
\end{abstract}

\section{Introduction}
Foundation models have achieved superior performance across various domains, including finance, healthcare, and cybersecurity. 
These models are trained on public data for general tasks and later fine-tuned by different industries for application-specific downstream tasks. 
The performance of these models relies heavily on the quality and diversity of their training data.
This is especially important for sensitive applications where the performance and robustness of the model are crucial.
However, for these applications, access to distributed and diverse data is often restricted through internal privacy policies or regulations such as HIPAA. 
This limitation is widely acknowledged as a primary barrier to training and fine-tuning models in the medical field \cite{liu2024moe}. 
Federated learning (FL) mitigates this issue by maintaining the locality of the data and only requiring the participating clients to share their local model (trained on their data) with a central server. 

While maintaining data locality can contribute to data privacy, recent attacks (e.g., \cite{balle2022reconstructing,hayes2024bounding}), demonstrated that updates shared with the server can leak significant information about the user dataset. 
These attacks can be categorized as follows, sorted by their privacy implications.
1) Membership Inference \cite{nasr2018machine, yeom2018privacy}: The attacker's goal is to determine whether a specific data point was part of the dataset used to train the model. 
2) Model inversion \cite{carlini2019secret}: A more severe privacy attack, where the attacker adversary attempts to recover information about the training data (i.e., approximate reconstruction). 
3) Training Data Extraction \cite{carlini2021extracting}: This is the most potent privacy attack where the attacker attempts to recover the original training samples precisely.

Differential privacy (DP) has long been the gold standard for mitigating these attacks \cite{dwork2006calibrating}. 
This is achieved by injecting a measured noise into model gradients to minimize the influence of any single data point on the model parameters. 
The magnitude of the noise is determined based on the desired privacy guarantee, measured by the privacy cost $(\epsilon,\delta)$, which serves as a key parameter in DP algorithms. 

In the FL setting, DP can be applied in different stages and by different parties to protect against data privacy attacks.
Depending on the privacy goals, often dictated by privacy policies or regulation requirements (e.g., HIPAA), DP can be applied by the clients adding noise to their gradient before sending it to the central server; this is often referred to as local DP (LDP). Alternatively, in central DP (CDP), the central server injects noise into the aggregated global model at the end of each training iteration. 
In CDP, the server has access to gradients sent by the users and is, therefore, assumed to be trusted.
 However, in applications that involve sensitive user data, relying on a trusted server can pose significant privacy risks and potentially violate privacy policies and regulations \cite{groce2011limits}.
 LDP can provide privacy at different stages of the model's life cycle (from model training to deployment) without assuming a trusted server. 

DP methods provide a worst-case privacy guarantee, which could result in utility loss due to the excessive added noise.  
In practice, the worst-case DP guarantees may not always be necessary for certain applications, potentially sacrificing performance by overprotecting the model.
For instance, in some applications, user participation might be public (e.g., social media), and only the user data should be protected. 
In such cases, the privacy goal is to defend against data reconstruction attacks to protect user data, which often require significantly less noise compared to the noise needed to defend against membership inference attacks \cite{balle2022reconstructing}.  

Another critical aspect of privacy-preserving FL is the selection of the DP method and the training parameters.
In highly distributed FL applications (e.g., \cite{yang2018applied}), the choice of DP method and parameters can significantly impact user participation. 
Certain DP methods tend to overestimate the necessary noise or impose a high computational overhead, making them impractical for low-end devices, particularly those from underrepresented groups. 
For example, as recently highlighted in \cite{birrell2024differentially}, most of the widely adopted DP methods, such as Renyi Differential Privacy (\rdp) \cite{mironov2017renyi}, rely on Poisson subsampling which results in producing variable size minibatches. 
This directly affects the machine's memory usage (Figure~\ref{fig:memory}).
While this might be manageable in traditional centralized settings, where the model training happens on powerful servers, in distributed FL applications with low-end devices, this can lead to an out-of-memory (OOM) error, potentially preventing certain devices from participating in the training process. 
This can directly affect the representation of different user groups in the training process, impacting the data diversity and potentially the model fairness \cite{fu2023client,xiao2021vehicle}. 
Lastly, the selection of FL parameters and the AI models to be trained in the federated setting can significantly impact the performance and utility of the trained model and user participation rates. 
For example, while a larger batch size may reduce the noise required for differential privacy, it can also introduce additional performance overhead for users. 
These challenges highlight a key research gap.

To address this, in this work, we present a framework for Federated Learning Implementation with Privacy (\oursys) that integrates a privacy practitioner into the training process to guide privacy-aware decision-making.
The practitioner's role is to help guide the selection of these parameters based on application requirements, AI model, privacy objectives, system specifications, resource constraints, and user participation dynamics. 
This approach ensures a more informed, context-sensitive decision-making process, optimizing both privacy protection and overall system performance and leading to the training of robust AI models.
Figure~\ref{fig:flip} provides a high-level overview of \oursys.
%
%

\subsection{Our Contributions}
The contributions of our work as as follows. 
To our knowledge, our work is the first to empirically highlight the importance and effect of this selection process in the private FL training of AI models. 
Our contributions are as follows. 

\noindent$\bullet$ \textbf{Adoption of a Privacy Practitioner:} Our work is the first to empirically highlight the significance and impact of selecting DP and FL parameters in privacy-preserving federated learning (FL) for AI models. We introduce a novel framework where a privacy practitioner assists in tuning these parameters based on factors such as privacy requirements, performance trade-offs, computational constraints, and system specifications.

\noindent $\bullet$\textbf{Adopting a Fixed Mini-Batch DP Method:} \oursys~is the first privacy-preserving FL framework to: 1) highlight the side effects of variable mini-batch sizes in existing DP methods (e.g., \rdp) on FL model training, 2) adopt a fixed mini-batch approach (i.e., \sys) to ensure stable memory usage throughout training, and 3) empirically compare the impact of this choice on model accuracy.

\noindent $\bullet$ \textbf{Comprehensive Study on DP, FL Parameters, and Data Distribution:} To our knowledge, \oursys~is the first framework to study the impacts of various DP and FL parameters and the data distributions among the users in FL settings on model performance. We conduct our experiments by fine-tuning Large Language Models (LLMs) on four well-known natural language processing tasks from the GLUE dataset \cite{glue_2019}. Fine-tuning LLMs in the context of studying the impact of DP on model performance is a well-established approach \cite{behnia2022ew,yu2021differentially}. For example, simulating our framework in fine-tuneing BERT (with 109M parameters) \cite{devlin2018bert} with two different target privacy costs, $\epsilon = 6$ and $\epsilon = 10$, while adopting \sys~as our DP method, incurs up to a 5\% accuracy loss compared to the non-private model. However, this gap can be reduced to as low as 2\% when optimized parameters and data distribution are enforced by the practitioner. Our framework is open-source for public verification and testing using the following link:

\begin{center}
    {\href{https://github.com/KasraAhmadi/FL-Privacy-LLM}{https://github.com/KasraAhmadi/FL-Privacy-LLM}}
\end{center}

 \section{Preliminaries}
In this section, we reivew
We examine three key areas of the literature: (1) FL, a distributed machine learning approach that trains models across multiple devices; (2) Differentially private deep learning, a comprehensive framework that ensures rigorous privacy during the learning process; and (3) The benefits of using differentially private stochastic gradient descent with fixed-size minibatches compared to Poisson-subsampled \rdp. 
\subsection{Federated Learning}

Federated Learning (FL) enables the distributed training of a central model, with contributions from a set of clients who each train a local copy of the model using their own data \cite{mcmahan2017communication}. 
FL considers a central server who aggregates the updates shared by the clients. 
 A generic FL system consists of  a central server and \( k \) clients. Each client \( C_i \) holds a local dataset \( D_i \), where \( i \in \{1, 2, \dots, k\} \). The server's objective is to train a model using data distributed across the \( k \) clients. When a client actively participates in local training, it aims to optimize a vector $\mathbf{w}$ for an AI model by minimizing a specified loss function. The server then aggregates the model weights received from the \( k \) clients as follows:\[ \mathrm{w}=\sum_{i=1}^{k} p_i\mathrm{w}_i \]
Here, \( \mathrm{w}_i \) represents the parameter vector trained by the \( i \)-th client, and  w is the aggregated parameter vector at the server. \( k \) denotes the total number of clients, while \( p_i = \frac{|D_i|}{|D|} \geq 0 \) satisfies \( \sum_{i=1}^{k} p_i = 1 \), with \( |D| = \sum_{i=1}^{k} |D_i| \) being the total number of data samples across all clients. This optimization problem can be expressed as:
\[ \mathrm{w^*}=\mathrm{argmin_w}\sum_{i=1}^{k}p_iF_i(\mathrm{w}) \] 
where \( F_i(w) \) represents the local loss function for the \( i \)-th client.

In the FL process, the \( k \) clients work together to train a machine-learning model with the assistance of a server. After several rounds of local training and updates exchanged between the server and the clients, the solution to the optimization problem is expected to converge to the globally optimal learning model.
\subsection{Differential Privacy}
Differential privacy \cite{dwork2006calibrating,dwork2014algorithmic} is a rigorous privacy framework that effectively mitigates the privacy risks associated with deep learning \cite{abadi2016deep}. The primary distinction between DP-based deep learning and standard deep learning lies in whether the gradient is released with privacy guarantees.\\
\textit{Definition 1}: A randomized algorithm  \( M \) is \((\epsilon, \delta)\)-differentially private if, for any two neighboring datasets \( S \) and \( S' \) (i.e., \( S' \) can be obtained by adding or removing a single data point from \( S \)), and for any event \( E \), the following condition holds :
\[ P[M(S) \in E] \leq e^\epsilon P[M(S') \in E] + \delta. \] 
We consider the \((\epsilon, \delta)\)-DP definition, where smaller values of \(\epsilon\) and \(\delta\) indicate a stronger privacy guarantee.

Differentially Private Stochastic Gradient Descent (DP-SGD) \cite{abadi2016deep} ensures differential privacy by introducing noise during the training process of machine learning models. DP-SGD modifies the standard mini-batch SGD algorithm by adding two additional steps:
\begin{itemize}
    \item Gradient Clipping: For each per-example gradient \( g(x_i) \), where \( x_i \) is a data point in the selected mini-batch, clip the \( l_2 \)-norm to a predefined threshold \( C \):  
   \[
   g(x_i) \gets \frac{g(x_i)}{\max(1, \|g(x_i)\|_2 / C)}.
   \]
   \item Noise Addition: Add Gaussian noise to the aggregated gradient of the mini-batch, where \( L \) is the mini-batch size and \( \sigma \) is the noise scale.:
   \[
   g \gets \frac{1}{L} \left( \sum_{i} g(x_i) + \mathcal{N}(0, \sigma^2 C^2) \right),
   \]
\end{itemize}
\subsection{\rdp~and \sys}
\begin{figure}[t]
    \centering
    \includegraphics[width=0.75\columnwidth]{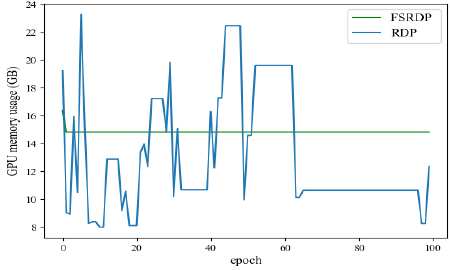}
    \caption{Comparison of \sys~and \rdp~accountant memory usage with a batch size of 120 and a dataset size of 50,000 per training epoch \cite{birrell2024differentially}.}
    \label{fig:memory}
\end{figure}
DP-SGD enables the use of a technique known as the moments accountant to sequentially monitor privacy leakage. This approach is encompassed by Rényi Differential Privacy (\rdp) \cite{mironov2017renyi}, a relaxed version of standard Differential Privacy \cite{dwork2016concentrated}. \rdp~is widely applied in private deep learning and is incorporated into modern DP libraries like Opacus\cite{opacus}. The underlying computation in \rdp~relies on subsampling techniques that use a privacy amplification lemma to enhance the privacy guarantees provided by the added noise.

While there have been previous attempts to compute privacy costs with a fixed mini-batch size, such as the works of Balle et al. \cite {balle2018privacy} for $(\epsilon,\delta)$-DP and Wang et al. \cite{wang2019subsampled} for \rdp, these approaches had significant shortcomings. The earlier $(\epsilon,\delta)$-DP methods did not compose easily over multiple training steps, often leading to privacy leakage, making them impractical for iterative processes like SGD. Similarly, Wang’s \rdp~accountant was not as tight, resulting in suboptimal privacy bounds. In contrast, \sys~\cite{birrell2024differentially} is the first privacy accountant capable of computing privacy costs with fixed-size mini-batches while achieving much tighter bounds. In fact, \sys~is very close to the theoretical lower bound in many practical cases, offering significantly improved privacy guarantees over previous RDP-based methods.

This offers a significant advantage of consistent memory usage compared to the variable-sized mini-batches in Poisson subsampling.
While the results in \cite{birrell2024differentially} were purely theoretical, in this paper, by highlighting the importance of fixed memory usage in FL settings, we adopt their accountant, and after conducting extensive experiments, we show that for certain applications and data distributions, the accuracy loss, compared to \rdp~is insignificant. 
Figure 1 depicts the memory consumption of \sys~and \rdp~accountants. In contrast to \rdp, \sys~maintains a constant memory footprint throughout the training process.

\section{Proposed Framework}
The goal of our framework, \oursys, is to optimize model performance given the required privacy requirements. 
In our proposed framework, the privacy practitioner (human expertise) is crucial in improving security and privacy through collaboration with clients and the privacy engine. The suggested data flow architecture, as depicted in Figure 2, includes four entities: Requirement, Privacy Practitioner, Clients, and Privacy Engine.

\subsection{Clients}
A client is a device, user, or entity involved in the decentralized training process, where it locally stores and processes its own private data. Clients contribute to training a shared machine learning model using their private data and periodically transmit model updates to the privacy engine. Since in \oursys~we aim to achieve full-stack privacy (during training and after deployment),  DP is implemented on the client side. 
Therefore, after receiving the FL and DP parameters from the practitioner, aside from model training, the client has to compute and inject the noise to its update. 
\subsection{Requirements}
In Federated Learning (FL), clients collaboratively train a centralized model, which may either be deployed for their own use or commercialized by the entity that organizes the federation. 
This entity facilitates client participation, often in exchange for a service or financial compensation.
Therefore, the selection of requirements is determined either by a coalition of clients or by the organizing entity overseeing the federation.
These requirements can be categorized into privacy and learning process requirements.
\subsubsection{Privacy Requirements}
\begin{itemize}
\item \textbf{Target privacy requirement:} This key parameter is central to calculating the privacy cost ($\epsilon$) and helps the practitioner determine the optimal trade-off between individual privacy and model accuracy. The selection of \(\epsilon\) varies by application, requiring a trade-off, as lower values of \(\epsilon\) enhance privacy but often come at the cost of reduced model accuracy. 

The interpretation of $\epsilon$ depends on many factors, including the number of training iterations (both local and global), data type, AI model, and other system specific parameters. However, this interpretation has proven to be very challenging \cite{pmlr-v119-triastcyn20a}, as $\epsilon$ does not directly translate to an intuitive measure of privacy risk across different settings . Instead, they can specify a privacy goal, such as mitigating the membership inference attack (MIA) or reconstruction attack, and a privacy practitioner can determine the suitable \(\epsilon\) value accordingly. 
Alternatively, if the DP is solely being implemented to meet certain requirements, enforced by regulations, the privacy cost could be computed by the privacy practitioner based on these requirements. 

MIA and reconstruction privacy goals are applied differently across various real-world scenarios. For example, in financial applications, MIA is generally not a primary concern. This is because certain information, such as the knowledge of which bank a person has a credit card with, is often considered public and not sensitive. However, more private details like one's Social Security Number (SSN) and account balance are crucial to protect. In such cases, the goal is not necessarily to prevent MIA but to guard against reconstruction attacks, where an adversary might attempt to reconstruct sensitive information about an individual from available data.
\end{itemize}
\begin{figure}[t]
    \centering
    \includegraphics[height=6cm, width=1\columnwidth]{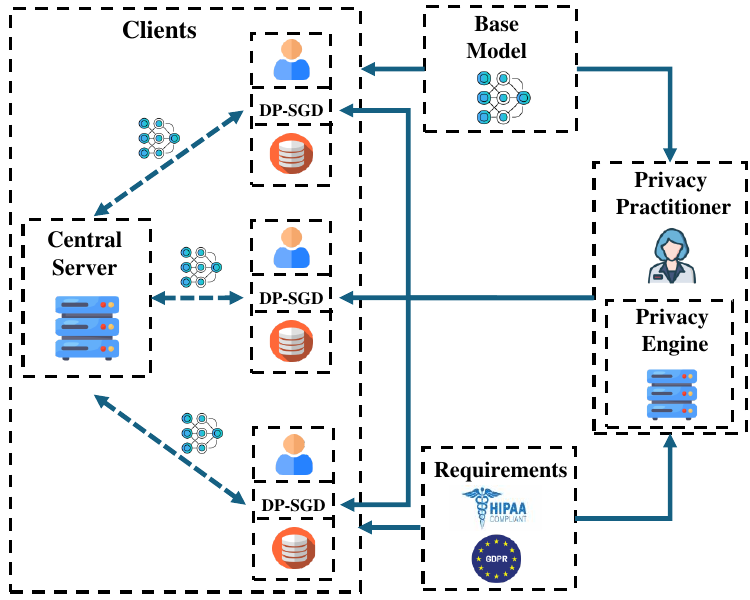}
    \caption{The data flow architecture for the proposed framework}
    \label{fig:flip}
\end{figure}
\subsubsection{Learning Process Requirements}
\begin{itemize}
    \item \textbf{Number of Clients:} The number of clients in the FL requires careful consideration. A larger client base increases diversity and robustness but necessitates efficient communication and aggregation strategies to address scalability issues. On the other hand, fewer clients make management simpler but may reduce the model's representativeness and resilience.
    \item \textbf{Data Distribution:}
   In some applications, no information about user data can be shared with the privacy practitioner. However, when feasible, insights into data distribution can significantly aid the practitioner in setting an effective privacy target and ultimately computing an appropriate $\epsilon$. As demonstrated in our experiments, when data is uniformly distributed across all users, the model generally achieves higher accuracy and can tolerate a lower $\epsilon$ while maintaining utility. Conversely, maintaining high accuracy under strict privacy constraints becomes more challenging in heterogeneous settings where data distributions vary significantly across clients. Certain data partitions may hold greater significance in the training process in such cases, requiring the privacy engine to incorporate adaptive strategies that account for these constraints, balancing privacy preservation with model performance.
   
   In addition to aiding in the selection of the appropriate $\epsilon$, knowledge of data distribution can also enable a tailored client selection during each training round. By carefully choosing clients with better uniform data distributions, the practitioner can improve overall data diversity across participants, model performance, and privacy. This tailored client selection approach can also help address other challenges posed by heterogeneous data distributions, ensuring a more balanced and effective training process in federated learning while maintaining privacy targets.
   
    \item \textbf{Client's Computation Constraints:} Computation constraints on the client side are another key concern in federated learning, as clients often consist of resource-limited devices such as smartphones, IoT devices, or edge devices. These limitations significantly impact the efficiency and feasibility of the learning process. In this context, we identify the benefit of the \sys~accountant \cite{birrell2024differentially}. Although not explicitly designed for resource-constrained devices, it is particularly valuable in our framework because it uses constant memory when calculating the noise for a given $\epsilon$, making it well-suited for these applications. Furthermore, the batch size, a parameter that directly influences both the training duration and the DP-SGD algorithm, is closely tied to the available memory on each client. It is the responsibility of the privacy practitioner to determine the batch size based on the memory capacity of individual clients (that can be provided in the requirements).  
\end{itemize}
\subsection{Privacy Practitioner}
\begin{itemize}
    \item \textbf{Security Parameters Calculation:}
   By utilizing its local privacy engine, the practitioner must determine the appropriate \(\epsilon\) value based on the target security constraints, which are influenced by the need to mitigate potential risks such as membership inference attacks (MIA) or reconstruction attacks, as well as to meet the privacy requirements set for the application. After selecting the suitable \(\epsilon\) value, the practitioner also chooses the appropriate differential privacy accountant. For memory-constrained applications, in our framework, \sys~ is selected, as it ensures constant memory usage but may result in slightly lower model accuracy. For cases where higher model accuracy is desired, \rdp~is chosen, which may involve non-constant memory usage. Once the \(\epsilon\) and \(\delta\) (where \(\delta = \frac{1}{|D_i|}\), representing the inverse of the dataset size) are determined, these parameters are provided clients. The client then utilizes their privacy engine, which calculates the necessary noise to be added to the client's gradient.
    \item \textbf{Set Batch size:}
    The batch size affects the efficiency and effectiveness of DP-SGD by determining how many samples are used in each model update. The ideal batch size strikes a balance between model security, training speed, memory usage, and overall performance. The privacy practitioner sets this parameter according to the available memory on the client side and the desired model performance.
\end{itemize}
\subsection{Privacy Engine}
\begin{itemize}
    \item \textbf{Noise Calculation:}
The privacy engine calculates the necessary noise using the specified accountant, \(\epsilon\), batch size, and \(\delta\) values. This noise is then injected into the client's gradient via the DP-SGD algorithm. 
\item \textbf{Requirement Adherence Tracking:}
In certain scenarios, the framework may fail to achieve the required accuracy due to factors such as the client's data partition, target security level, and target minimum accuracy. This limitation is primarily caused by the small size of partitioned data, where adding high noise to a gradient computed on a limited dataset can result in reduced accuracy. In such cases, the privacy engine generates a warning to indicate that the desired accuracy cannot be met. To address this, clients can either expand their data partitions or increase memory resources to adopt a different accountant, such as \rdp~ instead of \sys, to reduce the noise added during the DP-SGD algorithm.
\end{itemize}

\section{Experiments}
 
\subsection{Experiment Setup}
Our work was conducted using a single NVIDIA RTX 6000 Ada Generation GPU, equipped with 18,176 CUDA cores and 48GB of dedicated memory. We selected the pre-trained BERT base model (cased), which contains 109 million parameters, to ensure that LLM loading and fine-tuning could be accommodated within the internal memory. The model is available at Hugging Face hub\footnote{https://huggingface.co/google-bert/bert-base-cased}.
\\
We utilized the Flower framework\footnote{https://flowerai.net/} to simulate a federated learning environment for fine-tuning LLMs, while the Transformers library was used for tasks such as training, tokenization, and evaluation. The GLUE dataset\footnote{https://huggingface.co/datasets/nyu-mll/glue}, accessed via Hugging Face, was used for data loading. In particular, we utilized four specific datasets from the GLUE benchmark, which are described below.

\begin{itemize}
\item 
QNLI: The Question-Answering Natural Language Inference (QNLI) dataset, sourced from Wikipedia, comprises 110,400 question-paragraph pairs. Each paragraph contains only one sentence that answers the associated question. The task for the language model is to identify whether a given sentence contains the correct answer to the question.
\item 
QQP \cite{wang2019superglue}: The Quora Question Pairs (QQP) dataset contains more than 400,000 pairs of questions, each labeled to show whether the questions are semantically equivalent, meaning they are paraphrases of each other. The task for the language model is to determine if one question is a paraphrase of the other.
\item 
SST2 \cite{socher2013recursive}: The Stanford Sentiment Treebank (SST2) dataset consists of 68,800 sentences from movie reviews, each annotated with its sentiment. The task for the language model is to classify the sentiment of a given sentence as either positive or negative.
\end{itemize}

We defined the training and testing splits for each dataset as follows: QNLI with 105,000 samples for training and 5,460 for testing; QQP with 364,000 samples for training and 391,000 for testing; and SST-2 with 67,000 samples for training and 1,820 for testing.

We simulate a federated learning environment with 4 distinct clients, each assigned its own training and testing dataset. The setup includes 5 training rounds, a learning rate of \( 2e-5 \), and a batch size of 550. We adopt the FedAvg algorithm of McMahan et al. \cite{mcmahan2017communication}
 
To replicate real-world data generation across decentralized devices, we distribute the training and testing data among clients using 4 distinct partitioning strategies: Iid, Linear, Square, and Exponential.\\
In the Iid policy, the partitioner creates partitions by randomly and uniformly sampling data from the dataset. \\In the Linear policy, partitions are created such that the size of each partition is linearly proportional to its ID. The amount of data assigned to each client increases linearly with the partition ID. For example, if the IDs range from 1 to \( k\), the client with ID 1 receives 1 unit of data, client 2 receives 2 units, and so on, until client \( k\), which receives \( k\) units.\\
In the Squared policy, the data assigned to each client is proportional to the square of the partition ID. For instance, if the IDs range from 1 to \( k\), the client with ID 1 receives 1 unit of data, client 2 receives 4 units, and so on, up to client \( k\), which receives \(k^2\) units.\\
In the Exponential policy, the data allocation is based on the exponential value of the partition ID. For example, if the IDs range from 1 to M, the client with ID 1 receives \(e^1\) units of data, client 2 receives \(e^2\) units, and so on, up to client \( k\), which receives \(e^k\) units.
 
We assessed and pre-computed the necessary noise for the specified security parameters and data partition size to be added during the learning process, utilizing two widely-used state-of-the-art differential privacy accountants: Renyi Differential Privacy (\rdp) \cite{abadi2016deep,mironov2017renyi} and \sys~Accountant \cite{birrell2024differentially}. 
We developed a function in the Flower framework that incorporates noise at the client side during training. The noise is added after each round, scaled by the standard deviation divided by the batch size (550 in our experiments).
For security parameters, we set \( \epsilon = 10,6 \) and \( \delta = 1e-6 \) for larger datasets (e.g., QNLI, and QQP, each containing several hundred thousand samples) and \( \delta = 1e-5 \) for the smaller dataset (e.g., SST-2, with tens of thousands of samples). Clipping norm is set to 3 for all experiments.
\begin{table}[t]
\caption{Max Accuracy Across Datasets and Partition Policies Using \nonoise, \rdp, and \sys~Accountant for \(\epsilon\) = 6 and \(\epsilon\) = 10}
\resizebox{\columnwidth}{!}{%
\begin{tabular}{|c|c|c|cc|cc|}
\hline
\multirow{2}{*}{\textbf{Dataset}} &
  \multirow{2}{*}{\textbf{\begin{tabular}[c]{@{}c@{}}Partition \\ Policy\end{tabular}}} &
  \multirow{2}{*}{\textbf{\nonoise}} &
  \multicolumn{2}{c|}{\textbf{\(\epsilon\) = 10}} &
  \multicolumn{2}{c|}{\textbf{\(\epsilon\) = 6}} \\ \cline{4-7} 
                      &             &      & \multicolumn{1}{c|}{\textbf{\rdp}} & \textbf{\sys} & \multicolumn{1}{c|}{\textbf{\rdp}} & \textbf{\sys} \\ \hline
                      & Iid         & 88\% & \multicolumn{1}{c|}{87\%}         & 86\%           & \multicolumn{1}{c|}{87\%}         & 86\%           \\
\multirow{2}{*}{QQP}  & Linear      & 88\% & \multicolumn{1}{c|}{88\%}         & 86\%           & \multicolumn{1}{c|}{86\%}         & 85\%           \\
                      & Square      & 89\% & \multicolumn{1}{c|}{88\%}         & 85\%           & \multicolumn{1}{c|}{85\%}         & 84\%           \\
                      & Exponential & 89\% & \multicolumn{1}{c|}{87\%}         & 85\%           & \multicolumn{1}{c|}{88\%}         & 84\%           \\ \hline
                      & Iid         & 87\% & \multicolumn{1}{c|}{87\%}         & 86\%           & \multicolumn{1}{c|}{86\%}         & 85\%           \\
\multirow{2}{*}{QNLI} & Linear      & 88\% & \multicolumn{1}{c|}{88\%}         & 87\%           & \multicolumn{1}{c|}{87\%}         & 83\%           \\
                      & Square      & 88\% & \multicolumn{1}{c|}{88\%}         & 86\%           & \multicolumn{1}{c|}{86\%}         & 84\%           \\
                      & Exponential & 88\% & \multicolumn{1}{c|}{85\%}         & 84\%           & \multicolumn{1}{c|}{88\%}         & 83\%           \\ \hline
                      & Iid         & 91\% & \multicolumn{1}{c|}{90\%}         & 89\%           & \multicolumn{1}{c|}{90\%}         & 89\%           \\
\multirow{2}{*}{SST2} & Linear      & 90\% & \multicolumn{1}{c|}{90\%}         & 89\%           & \multicolumn{1}{c|}{89\%}         & 87\%           \\
                      & Square      & 92\% & \multicolumn{1}{c|}{90\%}         & 89\%           & \multicolumn{1}{c|}{89\%}         & 88\%           \\
                      & Exponential & 91\% & \multicolumn{1}{c|}{91\%}         & 90\%           & \multicolumn{1}{c|}{90\%}         & 89\%           \\ \hline
\end{tabular}%
}
\end{table}

\begin{table}[t]
\caption{Data Partition size based on Iid, Linear, Square, and Exponential partition Policies}
\resizebox{\columnwidth}{!}{%
\begin{tabular}{|c|c|c|c|c|c|}
\hline
\textbf{Dataset} &
  \textbf{\begin{tabular}[c]{@{}c@{}}Partition \\ Policy\end{tabular}} &
  \textbf{\begin{tabular}[c]{@{}c@{}}Partition\\ 1\end{tabular}} &
  \textbf{\begin{tabular}[c]{@{}c@{}}Partition\\ 2\end{tabular}} &
  \textbf{\begin{tabular}[c]{@{}c@{}}Partition\\ 3\end{tabular}} &
  \textbf{\begin{tabular}[c]{@{}c@{}}Partition\\ 4\end{tabular}} \\ \hline
                      & Iid         & 90962 & 90962 & 90962  & 90962  \\
\multirow{2}{*}{QQP}  & Linear      & 36384 & 72769 & 109153 & 145540 \\
                      & Square      & 12128 & 48512 & 109153 & 194053 \\
                      & Exponential & 11664 & 31707 & 86188  & 234287 \\ \hline
                      & Iid         & 26186 & 26186 & 26186  & 26185  \\
\multirow{2}{*}{QNLI} & Linear      & 10474 & 20948 & 31422  & 41899  \\
                      & Square      & 3491  & 13965 & 31422  & 55865  \\
                      & Exponential & 3357  & 9127  & 24811  & 67448  \\ \hline
                      & Iid         & 16838 & 16837 & 16837  & 16837  \\
\multirow{2}{*}{SST2} & Linear      & 6734  & 13469 & 20204  & 26942  \\
                      & Square      & 2244  & 8979  & 20204  & 35922  \\
                      & Exponential & 2159  & 5869  & 15953  & 43368  \\ \hline
\end{tabular}%
}
\end{table}

\begin{table*}[t]
\caption{Required Noise per Data Partition to Achieve \(\epsilon\) = 6 and \(\epsilon\) = 10 for accountants \rdp~and \sys }
\resizebox{\textwidth}{!}{%
\begin{tabular}{|c|c|cccccccc|cccccccc|}
\hline
\multirow{3}{*}{\textbf{Dataset}} &
  \multirow{3}{*}{\textbf{\begin{tabular}[c]{@{}c@{}}Partition \\ Policy\end{tabular}}} &
  \multicolumn{8}{c|}{\textbf{\(\epsilon\) = 10}} &
  \multicolumn{8}{c|}{\textbf{\(\epsilon\) = 6}} \\ \cline{3-18} 
 &
   &
  \multicolumn{2}{c|}{\textbf{Partition 1}} &
  \multicolumn{2}{c|}{\textbf{Partition 2}} &
  \multicolumn{2}{c|}{\textbf{Partition 3}} &
  \multicolumn{2}{c|}{\textbf{Partition 4}} &
  \multicolumn{2}{c|}{\textbf{Partition 1}} &
  \multicolumn{2}{c|}{\textbf{Partition 2}} &
  \multicolumn{2}{c|}{\textbf{Partition 3}} &
  \multicolumn{2}{c|}{\textbf{Partition 4}} \\ \cline{3-18} 
 &
   &
  \multicolumn{1}{c|}{\textbf{\rdp}} &
  \multicolumn{1}{c|}{\textbf{\sys}} &
  \multicolumn{1}{c|}{\textbf{\rdp}} &
  \multicolumn{1}{c|}{\textbf{\sys}} &
  \multicolumn{1}{c|}{\textbf{\rdp}} &
  \multicolumn{1}{c|}{\textbf{\sys}} &
  \multicolumn{1}{c|}{\textbf{\rdp}} &
  \textbf{\sys} &
  \multicolumn{1}{c|}{\textbf{\rdp}} &
  \multicolumn{1}{c|}{\textbf{\sys}} &
  \multicolumn{1}{c|}{\textbf{\rdp}} &
  \multicolumn{1}{c|}{\textbf{\sys}} &
  \multicolumn{1}{c|}{\textbf{\rdp}} &
  \multicolumn{1}{c|}{\textbf{\sys}} &
  \multicolumn{1}{c|}{\textbf{\rdp}} &
  \textbf{\sys} \\ \hline
 &
  Iid &
  \multicolumn{1}{c|}{0.84} &
  \multicolumn{1}{c|}{1.66} &
  \multicolumn{1}{c|}{0.84} &
  \multicolumn{1}{c|}{1.66} &
  \multicolumn{1}{c|}{0.84} &
  \multicolumn{1}{c|}{1.66} &
  \multicolumn{1}{c|}{0.84} &
  1.66 &
  \multicolumn{1}{c|}{0.91} &
  \multicolumn{1}{c|}{1.79} &
  \multicolumn{1}{c|}{0.91} &
  \multicolumn{1}{c|}{1.79} &
  \multicolumn{1}{c|}{0.91} &
  \multicolumn{1}{c|}{1.79} &
  \multicolumn{1}{c|}{0.91} &
  1.79 \\
QQP &
  Linear &
  \multicolumn{1}{c|}{0.96} &
  \multicolumn{1}{c|}{1.88} &
  \multicolumn{1}{c|}{0.86} &
  \multicolumn{1}{c|}{1.71} &
  \multicolumn{1}{c|}{0.82} &
  \multicolumn{1}{c|}{1.63} &
  \multicolumn{1}{c|}{0.8} &
  1.59 &
  \multicolumn{1}{c|}{1.12} &
  \multicolumn{1}{c|}{2.09} &
  \multicolumn{1}{c|}{0.95} &
  \multicolumn{1}{c|}{1.85} &
  \multicolumn{1}{c|}{0.89} &
  \multicolumn{1}{c|}{1.75} &
  \multicolumn{1}{c|}{0.86} &
  1.69 \\
 &
  Square &
  \multicolumn{1}{c|}{1.28} &
  \multicolumn{1}{c|}{2.39} &
  \multicolumn{1}{c|}{0.91} &
  \multicolumn{1}{c|}{1.8} &
  \multicolumn{1}{c|}{0.82} &
  \multicolumn{1}{c|}{1.63} &
  \multicolumn{1}{c|}{0.77} &
  1.55 &
  \multicolumn{1}{c|}{1.65} &
  \multicolumn{1}{c|}{2.84} &
  \multicolumn{1}{c|}{1.03} &
  \multicolumn{1}{c|}{1.98} &
  \multicolumn{1}{c|}{0.89} &
  \multicolumn{1}{c|}{1.75} &
  \multicolumn{1}{c|}{0.83} &
  1.63 \\
 &
  Exponential &
  \multicolumn{1}{c|}{1.29} &
  \multicolumn{1}{c|}{2.42} &
  \multicolumn{1}{c|}{0.98} &
  \multicolumn{1}{c|}{1.92} &
  \multicolumn{1}{c|}{0.84} &
  \multicolumn{1}{c|}{1.67} &
  \multicolumn{1}{c|}{0.76} &
  1.52 &
  \multicolumn{1}{c|}{1.68} &
  \multicolumn{1}{c|}{2.88} &
  \multicolumn{1}{c|}{1.16} &
  \multicolumn{1}{c|}{2.16} &
  \multicolumn{1}{c|}{0.92} &
  \multicolumn{1}{c|}{1.81} &
  \multicolumn{1}{c|}{0.81} &
  1.6 \\ \hline
 &
  Iid &
  \multicolumn{1}{c|}{1.02} &
  \multicolumn{1}{c|}{1.99} &
  \multicolumn{1}{c|}{1.02} &
  \multicolumn{1}{c|}{1.99} &
  \multicolumn{1}{c|}{1.02} &
  \multicolumn{1}{c|}{1.99} &
  \multicolumn{1}{c|}{1.02} &
  1.99 &
  \multicolumn{1}{c|}{1.24} &
  \multicolumn{1}{c|}{2.26} &
  \multicolumn{1}{c|}{1.24} &
  \multicolumn{1}{c|}{2.26} &
  \multicolumn{1}{c|}{1.24} &
  \multicolumn{1}{c|}{2.26} &
  \multicolumn{1}{c|}{1.24} &
  2.26 \\
\multirow{2}{*}{QNLI} &
  Linear &
  \multicolumn{1}{c|}{1.34} &
  \multicolumn{1}{c|}{2.49} &
  \multicolumn{1}{c|}{1.08} &
  \multicolumn{1}{c|}{2.09} &
  \multicolumn{1}{c|}{0.98} &
  \multicolumn{1}{c|}{1.92} &
  \multicolumn{1}{c|}{0.93} &
  1.84 &
  \multicolumn{1}{c|}{1.76} &
  \multicolumn{1}{c|}{2.99} &
  \multicolumn{1}{c|}{1.34} &
  \multicolumn{1}{c|}{2.4} &
  \multicolumn{1}{c|}{1.17} &
  \multicolumn{1}{c|}{2.17} &
  \multicolumn{1}{c|}{1.07} &
  2.03 \\
 &
  Square &
  \multicolumn{1}{c|}{2.13} &
  \multicolumn{1}{c|}{3.77} &
  \multicolumn{1}{c|}{1.22} &
  \multicolumn{1}{c|}{2.3} &
  \multicolumn{1}{c|}{0.98} &
  \multicolumn{1}{c|}{1.92} &
  \multicolumn{1}{c|}{0.89} &
  1.76 &
  \multicolumn{1}{c|}{2.93} &
  \multicolumn{1}{c|}{4.8} &
  \multicolumn{1}{c|}{1.56} &
  \multicolumn{1}{c|}{2.71} &
  \multicolumn{1}{c|}{1.17} &
  \multicolumn{1}{c|}{2.17} &
  \multicolumn{1}{c|}{1} &
  1.93 \\
 &
  Exponential &
  \multicolumn{1}{c|}{2.17} &
  \multicolumn{1}{c|}{3.85} &
  \multicolumn{1}{c|}{1.41} &
  \multicolumn{1}{c|}{2.6} &
  \multicolumn{1}{c|}{1.04} &
  \multicolumn{1}{c|}{2.01} &
  \multicolumn{1}{c|}{0.87} &
  1.72 &
  \multicolumn{1}{c|}{2.99} &
  \multicolumn{1}{c|}{4.9} &
  \multicolumn{1}{c|}{1.86} &
  \multicolumn{1}{c|}{3.15} &
  \multicolumn{1}{c|}{1.26} &
  \multicolumn{1}{c|}{2.3} &
  \multicolumn{1}{c|}{0.96} &
  1.87 \\ \hline
 &
  Iid &
  \multicolumn{1}{c|}{1.15} &
  \multicolumn{1}{c|}{2.19} &
  \multicolumn{1}{c|}{1.15} &
  \multicolumn{1}{c|}{2.19} &
  \multicolumn{1}{c|}{1.15} &
  \multicolumn{1}{c|}{2.19} &
  \multicolumn{1}{c|}{1.15} &
  2.19 &
  \multicolumn{1}{c|}{1.45} &
  \multicolumn{1}{c|}{2.56} &
  \multicolumn{1}{c|}{1.45} &
  \multicolumn{1}{c|}{2.56} &
  \multicolumn{1}{c|}{1.45} &
  \multicolumn{1}{c|}{2.56} &
  \multicolumn{1}{c|}{1.45} &
  2.56 \\
\multirow{2}{*}{SST2} &
  Linear &
  \multicolumn{1}{c|}{1.58} &
  \multicolumn{1}{c|}{2.88} &
  \multicolumn{1}{c|}{1.23} &
  \multicolumn{1}{c|}{2.32} &
  \multicolumn{1}{c|}{1.09} &
  \multicolumn{1}{c|}{2.1} &
  \multicolumn{1}{c|}{1.02} &
  1.98 &
  \multicolumn{1}{c|}{2.13} &
  \multicolumn{1}{c|}{3.54} &
  \multicolumn{1}{c|}{1.58} &
  \multicolumn{1}{c|}{2.75} &
  \multicolumn{1}{c|}{1.36} &
  \multicolumn{1}{c|}{2.43} &
  \multicolumn{1}{c|}{1.23} &
  2.25 \\
 &
  Square &
  \multicolumn{1}{c|}{2.7} &
  \multicolumn{1}{c|}{4.79} &
  \multicolumn{1}{c|}{1.42} &
  \multicolumn{1}{c|}{2.61} &
  \multicolumn{1}{c|}{1.09} &
  \multicolumn{1}{c|}{2.1} &
  \multicolumn{1}{c|}{0.96} &
  1.88 &
  \multicolumn{1}{c|}{3.75} &
  \multicolumn{1}{c|}{6.27} &
  \multicolumn{1}{c|}{1.88} &
  \multicolumn{1}{c|}{3.17} &
  \multicolumn{1}{c|}{1.36} &
  \multicolumn{1}{c|}{2.43} &
  \multicolumn{1}{c|}{1.12} &
  2.1 \\
 &
  Exponential &
  \multicolumn{1}{c|}{2.77} &
  \multicolumn{1}{c|}{4.9} &
  \multicolumn{1}{c|}{1.68} &
  \multicolumn{1}{c|}{3.03} &
  \multicolumn{1}{c|}{1.17} &
  \multicolumn{1}{c|}{2.22} &
  \multicolumn{1}{c|}{0.93} &
  1.83 &
  \multicolumn{1}{c|}{3.84} &
  \multicolumn{1}{c|}{6.45} &
  \multicolumn{1}{c|}{2.27} &
  \multicolumn{1}{c|}{3.74} &
  \multicolumn{1}{c|}{1.48} &
  \multicolumn{1}{c|}{2.6} &
  \multicolumn{1}{c|}{1.06} &
  2.02 \\ \hline
\end{tabular}%
}
\end{table*}
\subsection{Experiments Results}
Figures 3, 4, and 5 depict the accuracy results for three noise addition methods—\nonoise, \sys, and \rdp—across various datasets and partitioning policies in 5 rounds of the federated learning process. We examined the noise levels needed for each accountant to achieve a target \( \epsilon \) and evaluated how different partitioning policies influence the maximum accuracy attained.
\subsubsection{Various Partition Policies Impact on Max Accuracy}Table I presents the maximum accuracy achieved across various datasets, partition policies, \( \epsilon\) values, and accountant methods.
For small datasets like SST2, different partitioning policies have minimal impact on accuracy for both \rdp~and \sys.  
For large datasets without noise, partitioning policies similarly show little effect on accuracy.  
When dealing with large datasets, high security (\( \epsilon\) = 6), and the \sys~accountant, the Iid policy delivers the best performance.  
In contrast, for the same security level and large datasets using the \rdp~accountant, the Exponential policy performs best.  
For large datasets requiring lower accuracy (\( \epsilon\) = 10) with both \sys~and \rdp~accountants, partitioning policies have a negligible impact on performance.
Table II describes data partition size based on various partition policies. Finally, Table III outlines the necessary noise standard deviation for \( \epsilon = 10 \) and \( \epsilon = 6 \), which must be added during the training phase of each client at the end of each training round.
\subsubsection{Accountant Type Impact on Noise for Target \( \epsilon\)}In reference to Table III, achieving a higher privacy level (lower \( \epsilon\)) necessitates the addition of more noise. Furthermore, the \sys~accountant requires greater noise levels compared to the \rdp~accountant to attain the same epsilon value  under the add-remove adjacency relation. We note, however, that under replace-one adjacency, another commonly used adjacency notion, \rdp~and \sys~require nearly identical noise levels to achieve the same $\epsilon$ value; see the discussion in Section 4 of \cite{birrell2024differentially}, including the comparison in Figure 4. More specifically, the noise levels for \rdp~in Table III would need to be approximately doubled in order to provide the same $\epsilon$ guarantee under both adjacency relations, while the  noises currently listed for \sys~in Table III guarantee the stated $\epsilon$ for both adjacency relations. Thus, when requiring the privacy guarantees to extend to replace-one adjacency, the benefits of fixed-size subsampling are even more apparent.  As the proper choice of adjacency relation is debatable, one might reasonably require guarantees that cover both, in which case the benefits of \sys~are even more apparent.

\begin{figure}[t]
    \centering
    \includegraphics[width=1\columnwidth]{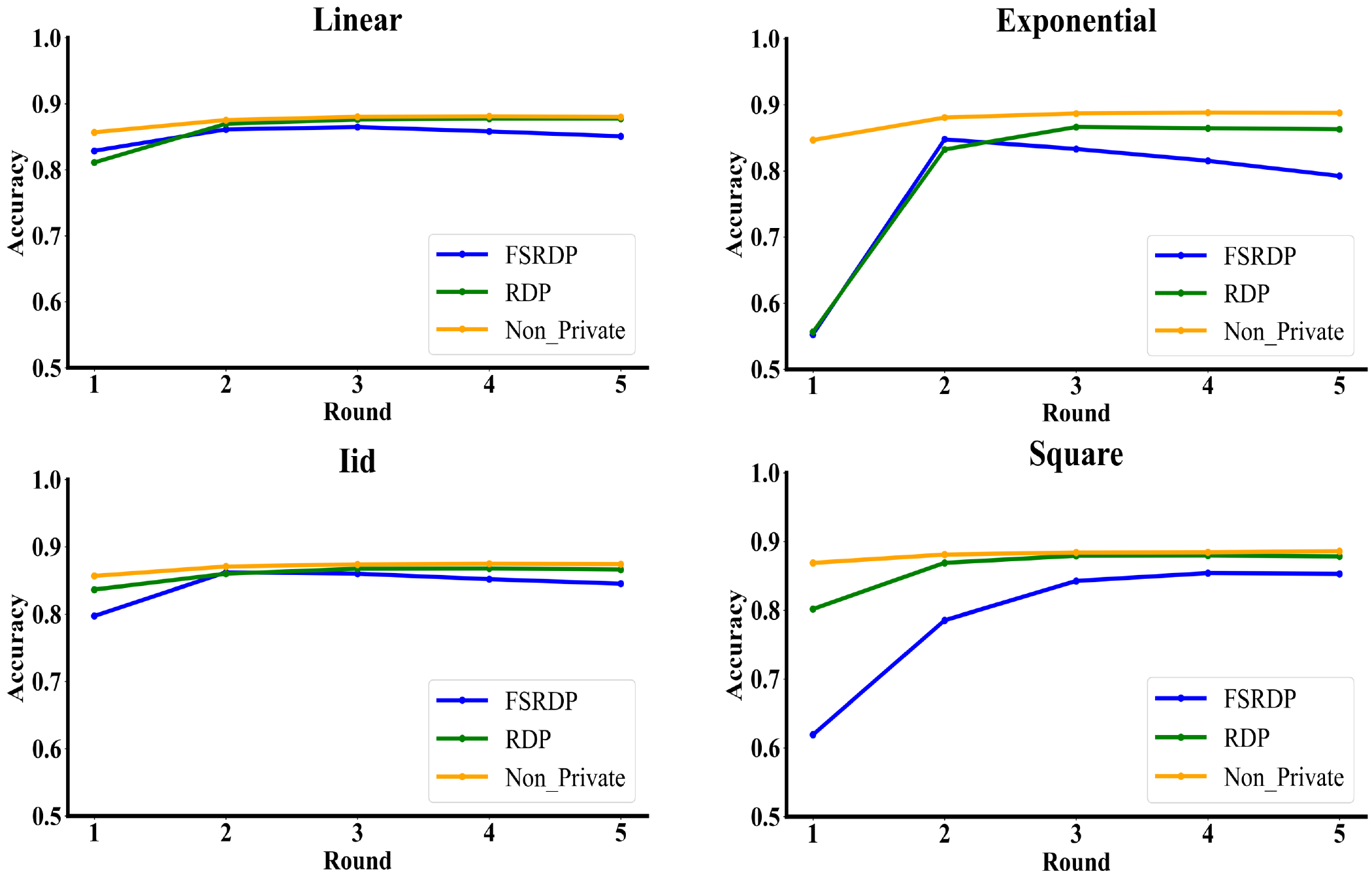}
    \caption{QQP Accuracy Across 5 Rounds with \( \epsilon\) = 10 }
    \label{fig:enter-label}
\end{figure}
\begin{figure}[t]
    \centering
    \includegraphics[width=1\columnwidth]{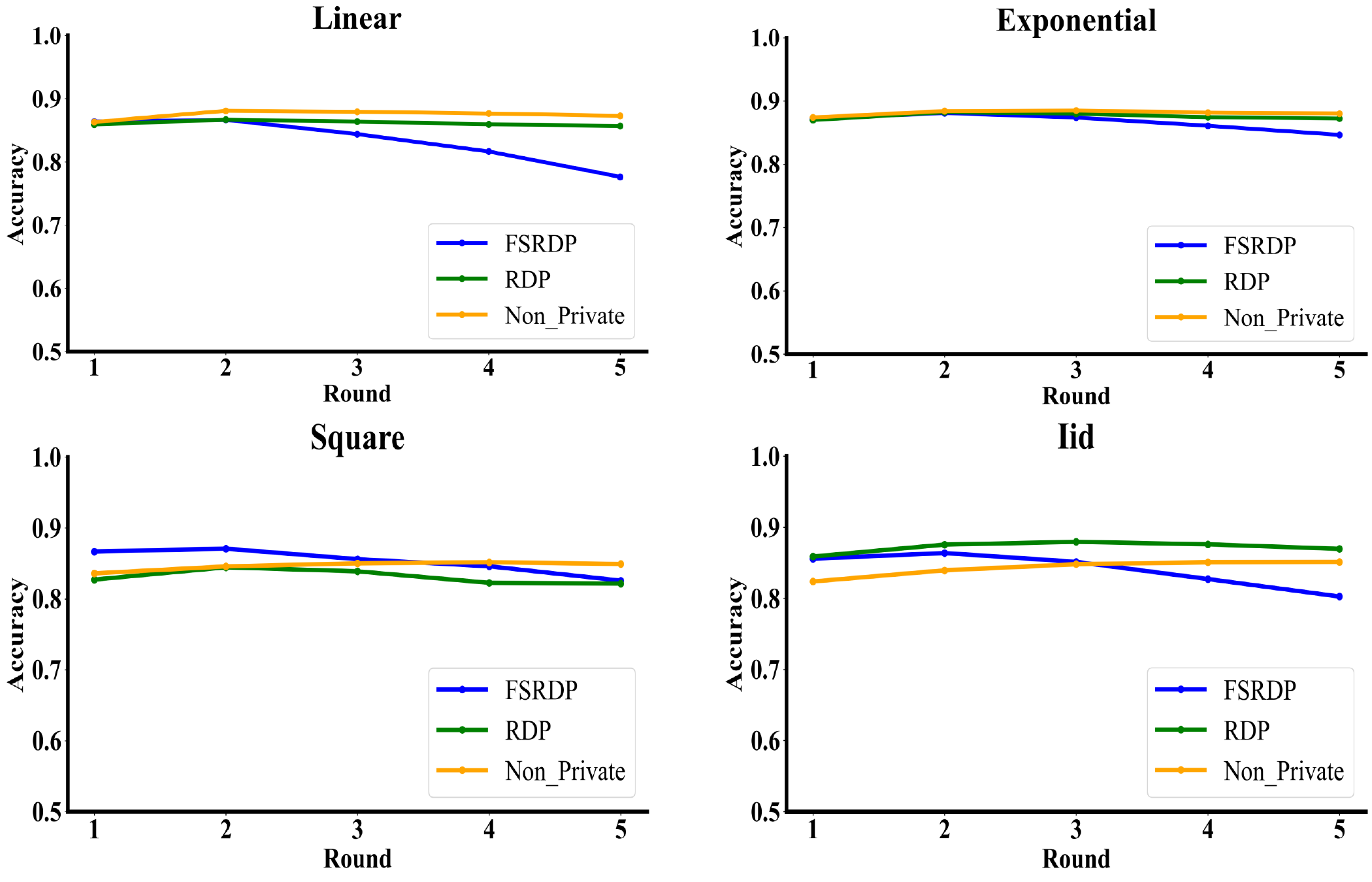}
    \caption{QNLI Accuracy Across 5 Rounds with \( \epsilon\) = 10 }
    \label{fig:enter-label}
\end{figure}
\begin{figure}[t]
    \centering
    \includegraphics[width=1\columnwidth]{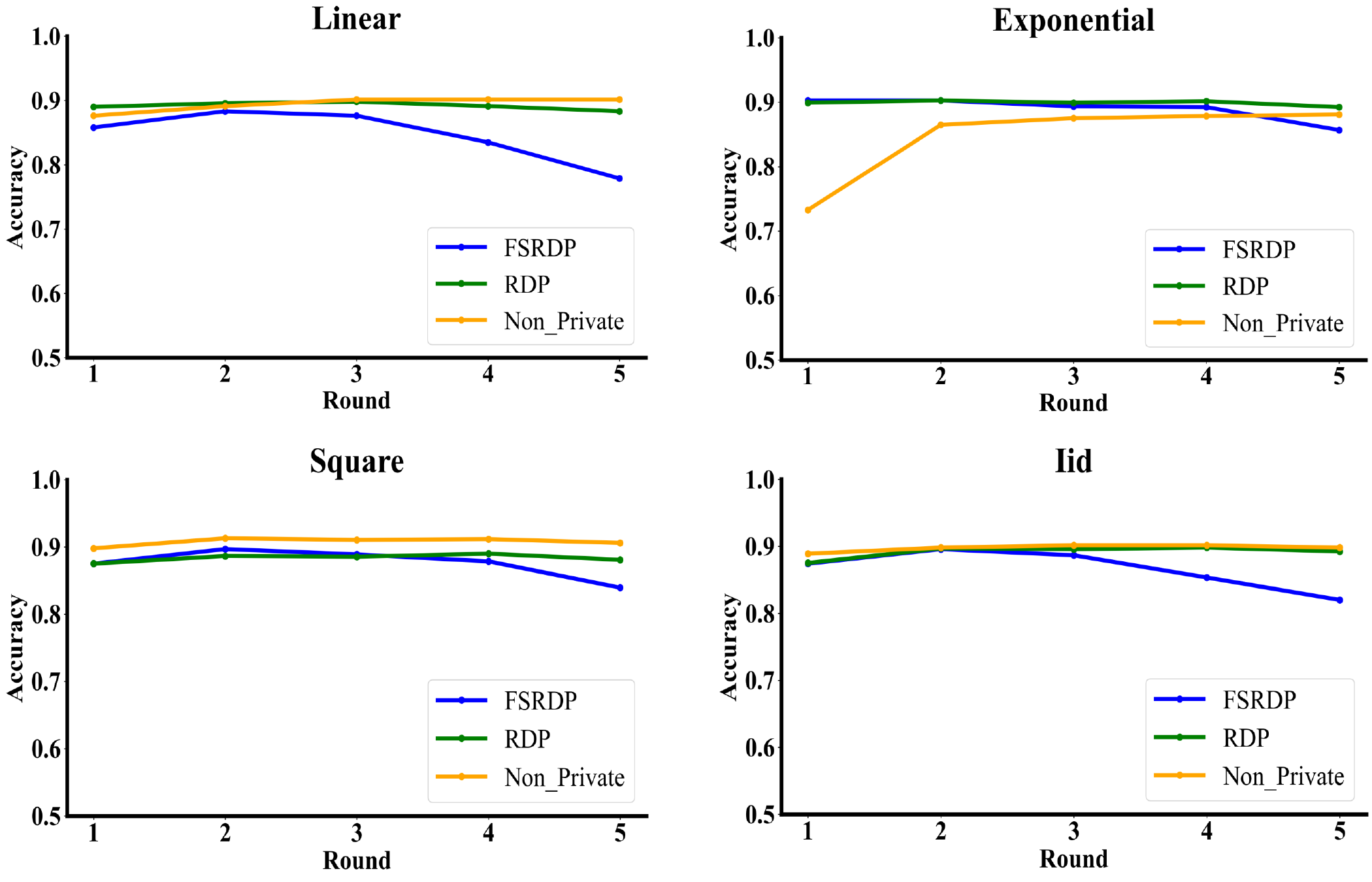}
    \caption{SST2 Accuracy Across 5 Rounds with \( \epsilon\) = 10 }
    \label{fig:enter-label}
\end{figure}

\subsection{Discussion}
Our research is the first in FL-DP to highlight the critical role of federated learning and differential privacy parameters, as well as their combined effect on model privacy and utility. 
\oursys~integrates a human practitioner who suggests privacy and FL parameters to help strike an optimal balance between privacy and utility in these environments. 
We focus on the widely adopted task of fine-tuning large language models (LLMs) and illustrate how key parameters such as privacy cost, data distribution, and client selection strategies affect model performance. 
\oursys~is also the first privacy-preserving framework to address the memory constraints of mobile devices in an FL setting by employing a privacy accountant with fixed memory requirements, achieved through a fixed minibatch size. 
Finally, we offer a detailed comparison between fixed-size minibatch accounting and the state-of-the-art \rdp~approach, highlighting the trade-offs in privacy and utility. 
The \sys~accountant provides advantages such as achieving an acceptable maximum accuracy and uniform memory usage. 
However, our experiments indicate that while the model performs well in the initial rounds using \sys~accountant, its accuracy may decline in the later stages.
This drop is caused by the cumulative effect of noise introduced by the \sys~accountant and imbalances in client contributions.
To overcome these challenges, we propose strategies like dynamically adjusting noise levels during training to better balance privacy and accuracy, as well as ensuring balanced client sampling to improve stability in the later rounds.
\section{Conclusion}
Our study highlights the critical role of parameter selection and the interpretation of privacy costs in different application settings. 
By examining the interplay between federated learning and differential privacy, we demonstrate how thoughtful parameter tuning can significantly impact both model utility and privacy guarantees. 
A promising direction for future work is to expand this study with more comprehensive experiments, considering diverse data types such as images and text to further generalize our findings. 
Additionally, further evaluation of \sys~with the replace-one adjacency relation could provide deeper insights into its effect on privacy guarantees and model utility, offering valuable guidance for privacy-preserving federated learning deployments.

\bibliographystyle{ieeetr}
\bibliography{reference}

\end{document}